%% file: main.tex
\documentclass[10pt]{article}
\input{packages}
\input{preamble}

\title{SceneParser: Hierarchical Scene Parsing for Visual Semantics Understanding}

\input{author}

\begin{document}

\maketitle

\clearpage
\input{Sections/0_abs}
\input{Sections/1_intro}
\input{Sections/2_relate}
\input{Sections/3_method}

\input{Sections/4_exp}
\input{Sections/5_con}

\bibliographystyle{unsrtnat}
\bibliography{main}

\clearpage
\appendix
\input{Sections/X_app}

\end{document}

%% file: packages.tex
\usepackage{hmi-preprint}
\usepackage{amsmath}
\usepackage{url}
\usepackage{booktabs}
\usepackage{nicefrac}
\usepackage{microtype}
\usepackage{marvosym}
\definecolor{cvprblue}{rgb}{0.21,0.49,0.74}
\hypersetup{
    colorlinks=true, 
    linkcolor=cvprblue, 
    citecolor=cvprblue, 
    urlcolor=cvprblue,  
}

\definecolor{cvprblue}{rgb}{0.21,0.49,0.74}

\usepackage[utf8]{inputenc} 
\usepackage[T1]{fontenc}    
\usepackage{amsfonts}       
\usepackage{graphicx}
\usepackage{amsmath}
\usepackage{multirow}
\usepackage{array}
\usepackage{booktabs}
\usepackage{multirow}
\usepackage{tabularx}
\usepackage{caption}

%% file: preamble.tex
\preprintlogos{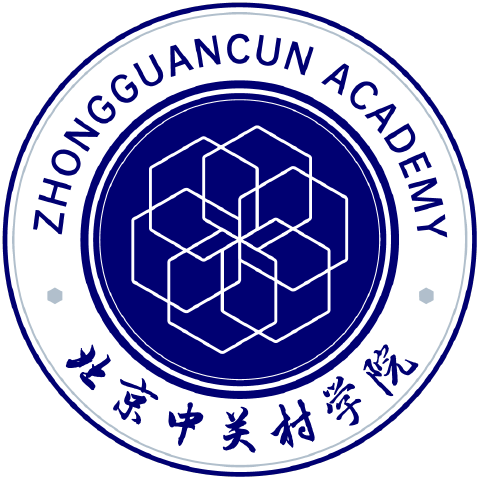}{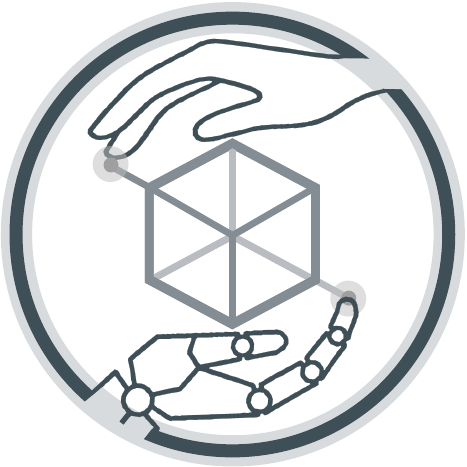}

\preprintdate{May 2026}

\runningtitle{SceneParser: Hierarchical Scene Parsing for Visual Semantics Understanding}

\preprintfooter{\textsuperscript{\Letter} Corresponding author.}

\preprintlinkset
  {https://github.com/ZGCA-HMI-Lab/SceneParser}
  {https://huggingface.co/SceneParser/SceneParser-model}
  {https://huggingface.co/datasets/SceneParser/SceneParser-bench}

\preprintteaser[fig:overview]{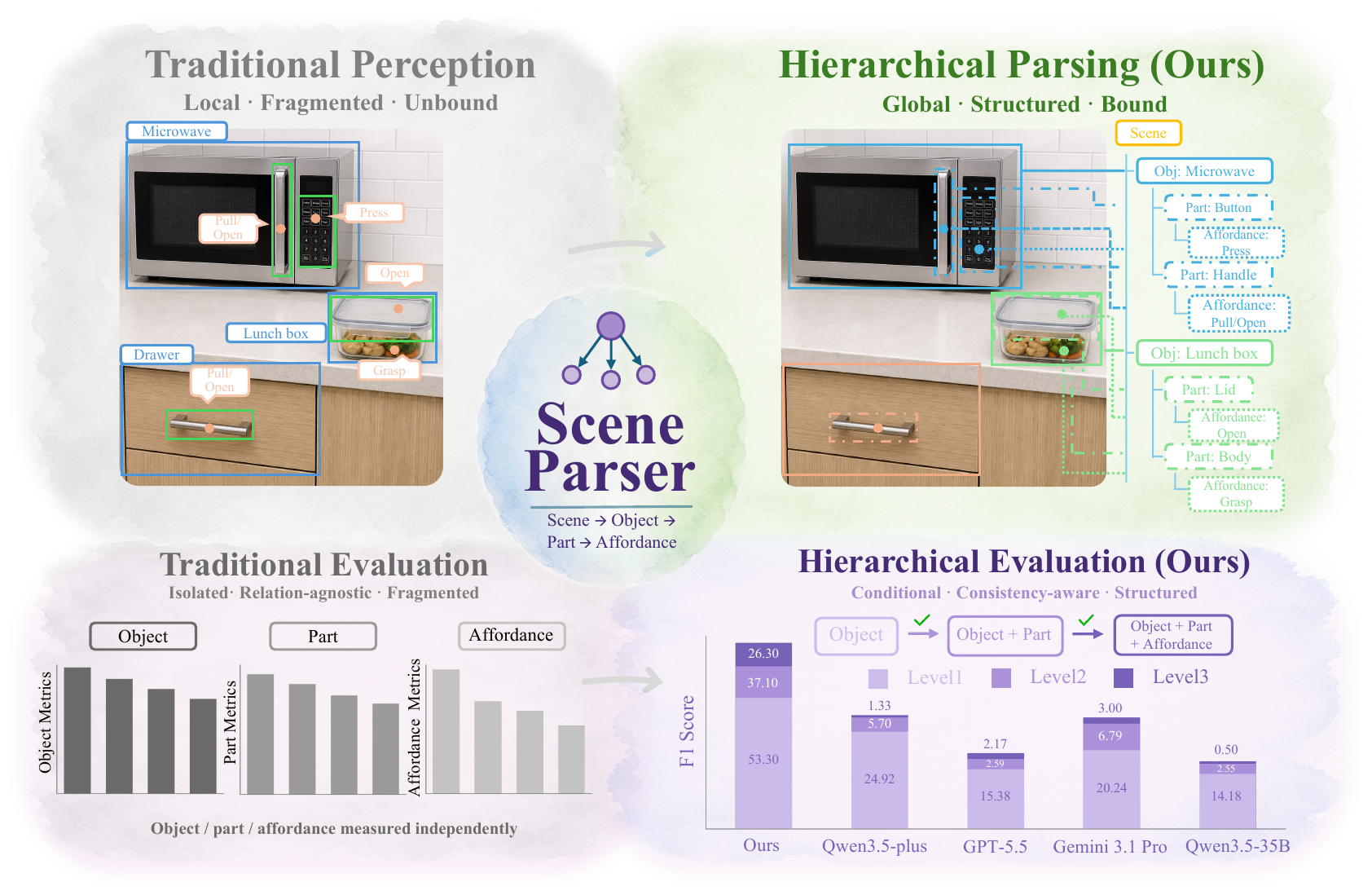}{SceneParser overview. 
    Conventional perception produces local predictions, such as object boxes, part regions, and affordance points, which are often evaluated independently and leave cross-level relations unverified. 
    In contrast, we parse a scene into an explicit \textit{scene} $\rightarrow$ \textit{object} $\rightarrow$ \textit{part} $\rightarrow$ \textit{affordance} hierarchy.
    Besides, our hierarchy-aware evaluation conditions lower-level correctness on upper-level matches, jointly assessing detection, cross-level binding, and structural consistency.}

\newcolumntype{C}[1]{>{\centering\arraybackslash}p{#1}}

\usepackage[most]{tcolorbox}
\tcbuselibrary{listings,breakable,skins}

\definecolor{schemaborder}{RGB}{165,190,220}
\definecolor{schemabg}{RGB}{248,252,255}
\definecolor{schematitlebg}{RGB}{228,240,253}

\newtcblisting{promptbox}[2][]{
  enhanced,
  breakable,
  colback=schemabg,
  colframe=schemaborder,
  colbacktitle=schematitlebg,
  coltitle=black!88,
  fonttitle=\normalsize\bfseries,
  title={#2},
  boxrule=0.4pt,
  arc=2.5pt,
  outer arc=2.5pt,
  titlerule=0pt,
  toptitle=3pt,
  bottomtitle=3pt,
  left=7pt,
  right=7pt,
  top=4pt,
  bottom=4pt,
  before skip=5pt,
  after skip=5pt,
  listing only,
  listing options={
    basicstyle=\ttfamily\scriptsize,
    breaklines=true,
    columns=fullflexible,
    keepspaces=true,
    showstringspaces=false
  },
  #1
}

%% file: author.tex
\author{
Pengxin Xu\textsuperscript{1,2} \quad
Xincheng Lin\textsuperscript{3} \quad
Luping Xiao\textsuperscript{2,4} \quad
Qing Jiang\textsuperscript{5} \quad
Meishan Zhang\textsuperscript{1} \quad
Hao Fei\textsuperscript{6\Letter}
\\
Shanghang Zhang\textsuperscript{7} \quad
Xingyu Chen\textsuperscript{2\Letter}
}

\affil{
\begin{tabular}{c}
\textsuperscript{1}Harbin Institute of Technology, Shenzhen \quad
\textsuperscript{2}Zhongguancun Academy \quad
\textsuperscript{3}Huazhong University of Science and Technology
\\
\textsuperscript{4}Beijing University of Posts and Telecommunications \quad
\textsuperscript{5}South China University of Technology
\\
\textsuperscript{6}University of Oxford \quad
\textsuperscript{7}Peking University
\end{tabular}
}

%% file: Sections/0_abs.tex
\begin{abstract}
General scene perception has progressed from object recognition toward open-vocabulary grounding, part localization, and affordance prediction. 
Yet these capabilities are often realized as isolated predictions that localize objects, parts, or interaction points without capturing the structured dependencies needed for interaction-oriented scene understanding.
To address this gap, 
we introduce \textit{Hierarchical Scene Parsing}, an interaction-oriented parsing task that represents physical scenes as explicit \textit{scene $\rightarrow$ object $\rightarrow$ part $\rightarrow$ affordance} hierarchies with cross-level bindings. We instantiate this task with SceneParser, a VLM-based parser trained for unified hierarchical generation with structural-completion pseudo labels and curriculum learning. To support training and evaluation, we construct SceneParser-Bench, a large-scale benchmark built with a scalable hierarchical data engine, containing 110K training images, a 5K validation split, 777K objects, 1.14M parts, 1.74M affordance annotations, and 1.74M valid object-part-affordance chain instances. We further introduce Level-1 to Level-3 conditional metrics and ParseRate to evaluate localization, cross-level binding, and hierarchical completeness. Experiments show that existing MLLMs and perception-stitching pipelines struggle with hierarchical parsing on our SceneParser-Bench, while SceneParser achieves stronger structure-aware performance. Besides, ablations, evaluations on COCO and AGD20K, and a downstream planning probe demonstrate that our SceneParser is compatible with conventional tasks and provides an actionable representation for visual understanding.
\end{abstract}

%% file: Sections/1_intro.tex
\section{Introduction}

Scene perception is a basic computer vision problem with wide downstream applications ~\citep{everingham2010pascal,lin2014microsoft,zhou2017scene}. For example, it is also a fundamental capability for embodied intelligence, supporting robotic navigation, manipulation, and human-robot interaction~\citep{brohan2024rt,majumdar2024openeqa,wang2024embodiedscan}. Recent progress has shifted visual understanding from closed-set object recognition toward open-world and language-conditioned localization~\citep{liu2024grounding,ren2024grounding,cheng2024yolo,xiao2024florence,fu2025llmdet,xiao2025towards}. Object grounding and referring expression comprehension enable models to localize targets specified by open-vocabulary categories, phrases, or natural-language instructions ~\citep{li2022grounded,liu2024grounding,chen2023shikra,zhang2024ferret,ma2024groma,jiang2024chatrex,jiang2025detect,wang2025vgr}. Beyond object-level grounding, embodied visual understanding further requires fine-grained and interaction-oriented perception, including part localization and affordance prediction. These advances suggest that modern visual perception is increasingly expected to localize action-relevant entities across multiple semantic levels~\citep{he2022partimagenet,ramanathan2023paco,luo2022learning,qian2024affordancellm,yuan2024robopoint,wang2026affordance,sun2025digital}.

However, multi-level localization does not necessarily amount to structured scene understanding. Existing methods may predict object boxes, part regions, and affordance points, but these outputs are often local, fragmented, and weakly bound across levels. An isolated affordance point, for example, does not explicitly encode the object it belongs to, the functional part that supports the action, or its relation to alternative interaction candidates; similarly, standard metrics often score objects, parts, and affordances independently, without verifying whether they form a consistent object-part-affordance chain. This limitation also aligns with a broader observation in agentic and spatial reasoning: downstream decision making often benefits from explicit intermediate representations, such as structured GUI elements~\citep{lu2024omniparser}, spatial layouts, entity registries, or visual primitives~\citep{hua2026unleashing,lu2026thinkingvisualprimitives}. For embodied physical scenes, however, the desired representation should be functional rather than merely spatial, describing objects, their actionable parts, and the affordances grounded on those parts.

Motivated by this perspective, we introduce \textbf{Hierarchical Scene Parsing}, an interaction-oriented parsing task for physical scenes, as illustrated in~\autoref{fig:overview}. Given an RGB scene image, the goal is to recover a hierarchy of \textit{scene $\rightarrow$ object $\rightarrow$ part $\rightarrow$ affordance}. Each object is represented by a semantic name and a bounding box; each part is represented by a part name and a part bounding box; and each affordance is represented by an action label and an interaction point. Crucially, the representation is not a flat collection of predictions: every part is explicitly attached to its parent object, and every affordance is grounded on a specific functional part. Therefore, instead of only asking where an object, part, or affordance point is located, hierarchical scene parsing asks which objects are present, how they decompose into functional parts, which parts support which actions, and where each action should be grounded.

To instantiate this task as a learnable model, we design \textbf{SceneParser}, a VLM-based hierarchical parser for unified object-part-affordance generation. Instead of stitching separate object, part, and affordance predictors, SceneParser directly generates a structured hierarchy with explicit parent-child bindings. To better adapt VLM generation to hierarchical targets, we introduce structural-completion pseudo labels to model valid non-expandable cases and reduce broken hierarchies. We further use curriculum learning to progressively balance object-level localization stability, part decomposition, affordance grounding, and hierarchical completeness.

To support training and evaluation, we construct \textbf{SceneParser-Bench}, a large-scale benchmark built with a scalable hierarchical data engine. The engine progressively constructs supervision at three levels: scene-level object grounding, object-centric part parsing, and affordance parsing with hierarchy reconstruction. Specifically, it grounds objects from generated names and referring expressions, parses functional parts from object-centric crops, and links generated affordance actions and interaction points to the corresponding object-part hierarchy through textual and geometric matching. SceneParser-Bench contains 110K training images and a 5K validation split, with validated annotations totaling 777K object instances, 1.14M part instances, 1.74M affordance annotations, and 1.74M valid object-part-affordance chain instances, covering 2,905 object categories, 12,349 part categories, and 85 affordance categories.

To evaluate hierarchical scene parsing, we introduce a structure-aware protocol with Level-1 to Level-3 conditional metrics and ParseRate. 
Level 1 evaluates object recognition and localization; Level 2 evaluates the object-part chain, counting a part as correct only when both the parent object and the part are correctly matched; and Level 3 evaluates the full object-part-affordance chain, counting an affordance as correct only when the object-part chain is valid, the action label is correct, and the interaction point is spatially valid. 
By conditioning lower-level correctness on upper-level matches, the protocol measures coherent hierarchical structures beyond independent object, part, and affordance scores. 
ParseRate further measures whether parse-eligible objects and scenes are structurally completed with the required part and affordance expansions.

Experiments show that hierarchical scene parsing remains challenging for existing MLLMs and perception-stitching pipelines: object-level localization does not reliably translate into valid part grounding, affordance grounding, or structural completeness. 
In contrast, SceneParser achieves stronger hierarchy-aware performance, and ablations show that nested object-part-affordance generation outperforms flat triplet prediction. 
Cross-task evaluations on COCO object detection and AGD20K affordance grounding further indicate transferability to conventional tasks. 
Finally, a downstream planning probe shows that providing the parsed scene hierarchy to a multimodal planner yields more complete plans and more consistent object-part-affordance selections than task-only prompting, supporting hierarchical scene parsing as a structured and actionable representation for embodied visual understanding.

Our contributions are summarized as follows:
\begin{itemize}
    \item We formulate Hierarchical Scene Parsing as an interaction-oriented parsing task for physical scenes, where visual understanding is defined as recovering explicit \textit{scene $\rightarrow$ object $\rightarrow$ part $\rightarrow$ affordance} structures with cross-level bindings.

    \item We construct SceneParser-Bench, a large-scale training data and benchmark built with a scalable hierarchical data engine, and introduce a structure-aware evaluation protocol with Level-1 to Level-3 conditional metrics and ParseRate to assess localization, cross-level binding, and hierarchical completeness.

    \item We propose SceneParser, a VLM-based hierarchical parser with unified object-part-affordance generation, structural-completion pseudo labels, and curriculum learning. Compared to existing MLLMs or detectors, we achieve stronger hierarchy-aware parsing on SceneParser-Bench and competitive results on conventional tasks of COCO and AGD20K. 
\end{itemize}

%% file: Sections/2_relate.tex
\section{Related Work}

\subsection{Object Grounding and referring}
Object grounding has progressed from closed-set object detection to open-vocabulary and language-conditioned localization. Classical detectors predict object boxes under predefined vocabularies, ranging from CNN-based models to Transformer-based detectors~\citep{redmon2016you,ren2015faster,carion2020end,zhang2022dino}. Open-vocabulary detectors, such as GLIP and Grounding DINO, further localize objects described by arbitrary categories, phrases, or referring expressions~\citep{li2022grounded,liu2024grounding,ren2024dino,cheng2024yolo,wang2025yoloe,jiang2024t,minderer2023scaling, cho2024language}. Recent MLLM-based grounding methods instead cast localization as coordinate generation under natural-language prompts~\citep{chen2021pix2seq,peng2023kosmos,chen2023shikra,you2023ferret,ma2024groma,jiang2024chatrex,jiang2025detect,rasheed2024glamm, lai2024lisa}. Despite strong object-level localization, these methods mainly output grounded object instances and do not explicitly decompose objects into functional parts or bind affordances to those parts. In our formulation, object grounding is only the first-level anchor for hierarchical scene parsing, followed by conditional part parsing and affordance grounding.

\subsection{Part and Affordance Understanding}
Beyond object grounding, fine-grained perception studies object parts and action-relevant regions. Part-level datasets such as PartImageNet, PACO, and ADE20K provide dense supervision for semantic or functional components grounding~\citep{he2022partimagenet,ramanathan2023paco,chen2014detect,zhou2017scene}, while open-vocabulary part segmentation extends part localization to flexible text queries~\citep{sun2023going,choi2024understanding,li2024partglee,wan2025instructpart}. These works show that parts are key intermediates, as many interactions depend on functional parts rather than whole objects.

Affordance understanding is more directly tied to embodied interaction. Early benchmarks such as UMD and AGD20K provide mask- or region-level supervision~\citep{myers2015affordance,nguyen2017object,luo2022learning}, while recent VLM- and robotics-oriented methods extend affordance grounding to open-world, instruction-conditioned, and reasoning-based settings~\citep{qian2024affordancellm,yuan2024robopoint,chen2025maskprompt,tang2025affordgrasp,wang2026affordance,hao2025roboafford++,ji2025robobrain,zhang2025a4,ma2025glover++,chu20253d,li2025learning,lee2025affogato}. However, most methods still formulate parts and affordances as local targets, such as masks, heatmaps, boxes, or points, leaving interaction structure implicit. 
Such outputs may localize action-relevant regions, but do not ensure that an affordance is supported by the correct functional part of the correct object. 
Our formulation makes these dependencies explicit through object-part-affordance chains, where affordance correctness is evaluated jointly with its parent object and part.



%% file: Sections/3_method.tex
\section{SceneParser-Bench}
\subsection{Problem Formulation}
\label{sec:Problem_Formulation}

We study \textit{hierarchical scene parsing} for embodied visual understanding. 
Given an RGB image $I$, the goal is to predict a structured scene representation that decomposes it into a \textit{scene $\rightarrow$ object $\rightarrow$ part $\rightarrow$ affordance} hierarchy. 
Unlike conventional perception outputs, the prediction must recover visual entities, locations, and parent-child bindings among objects, parts, and affordances.

Formally, we represent the parsed scene as a rooted hierarchy $\mathcal{H}=\{O_i\}_{i=1}^{N}$, where each object is $O_i=(c_i,b_i,\mathcal{P}_i)$. 
Here $c_i$ is the object name, $b_i=(x_1,y_1,x_2,y_2)$ is its bounding box, and $\mathcal{P}_i=\{P_{ij}\}_{j=1}^{M_i}$ denotes its parts. 
Each part is $P_{ij}=(q_{ij},b_{ij},\mathcal{A}_{ij})$, where $q_{ij}$ is the part name, $b_{ij}$ is the part bounding box, and $\mathcal{A}_{ij}=\{A_{ijk}\}_{k=1}^{K_{ij}}$ is the set of affordances grounded on this part. 
Each affordance is $A_{ijk}=(a_{ijk},u_{ijk})$, where $a_{ijk}$ is an action label and $u_{ijk}=(x,y)$ is the interaction point. 
Both $\mathcal{P}_i$ and $\mathcal{A}_{ij}$ may be empty, allowing the representation to model whether an object should be decomposed into parts and whether a part supports affordances.

This task is therefore not independent localization at each level. 
A part is valid only under its parent object, and an affordance is valid only under a valid object-part chain. 
The expected output is a serialized hierarchy $Y=\mathrm{Serialize}(\mathcal{H})$, not separate object, part, and affordance predictions.

\subsection{Evaluation Suite}
\label{sec:evaluation_suite}

We evaluate hierarchical scene parsing with a structure-aware protocol consisting of Level-1 to Level-3 conditional metrics and ParseRate. 
Following standard detection evaluation~\citep{everingham2010pascal,lin2014microsoft}, we report precision, recall, and F1, while enforcing hierarchical validity: parts are evaluated only under matched objects, and affordances only under matched object-part pairs.

\paragraph{Level-1 to Level-3 Metrics.}
Given an IoU threshold $\tau$, we define three conditional levels.

\textbf{Level 1 (Object)} measures object recognition and localization. 
Predicted and ground-truth objects are greedily matched one-to-one using identical object names and object-box IoU $\ge \tau$.

\textbf{Level 2 (Object-Part)} measures part parsing under matched objects. 
Within each matched object pair, predicted and ground-truth parts are greedily matched one-to-one using identical part names and part-box IoU $\ge \tau$. 
Thus, a part can be counted as correct only if its parent object is also correct.

\textbf{Level 3 (Object-Part-Affordance)} measures affordance grounding conditioned on a valid object-part chain. 
Within each matched object-part pair, predicted and ground-truth affordances are matched by identical action labels and spatial validity of the interaction point: the point must fall inside the ground-truth affordance box; if no affordance box is recorded, we use the parent part box as the fallback region. 
Thus, an affordance is correct only when its parent object, parent part, action label, and interaction point are all correct.

\paragraph{ParseRate.}

Level metrics evaluate the correctness of matched entities, but they do not directly measure whether a parse-eligible object is structurally completed. 
We therefore define ParseRate to measure hierarchical completeness over ground-truth objects that require part and/or affordance expansion. 
A matched parse-eligible object at IoU threshold $0.5$ is counted as complete if the prediction satisfies its required expansion pattern: parts for objects with parts, affordances for objects with affordances, and both when both are annotated.

Formally,
\[
\mathrm{ParseRate}
=
\frac{
\#\{\text{complete parse-eligible GT objects}\}
}{
\#\{\text{parse-eligible GT objects}\}
}.
\]
We report ParseRate at both object and scene levels. 
For scene-level ParseRate, counts are accumulated over all parse-eligible objects across scenes, providing a scene-centric measure of hierarchical coverage. 
Detailed definitions are provided in~\autoref{appendix:evaluation_details}.

\subsection{Data Engine}
\label{sec:data_engine}

To support hierarchical scene parsing, we build a three-stage data engine that converts a scene RGB images into structured \textit{scene $\rightarrow$ object $\rightarrow$ part $\rightarrow$ affordance} supervision, with the full pipeline shown in~\autoref{fig:data_engine}. 
It grounds scene-level objects, parses object-centric parts, extracts action-relevant affordance regions, and reconstructs them into unified hierarchies for structured generation training and hierarchy-aware evaluation.

\paragraph{Stage 1: Scene-level object grounding.}
Given a scene image, we use GPT-5~\citep{openai2025gpt5} to generate object names and referring expressions, which are then grounded by localization models, including Grounding DINO~\citep{liu2024grounding}, Rex-Omni~\citep{jiang2025detect}, and SAM3~\citep{carion2025sam3segmentconcepts}. 
We apply model-specific confidence thresholds and merge duplicated predictions across models to improve coverage and reduce redundant detections. 
The resulting object boxes serve as parent anchors for subsequent part and affordance parsing.

\paragraph{Stage 2: Object-centric part parsing.}
For each grounded object, we crop the corresponding image region to reduce background interference. 
GPT-5 proposes candidate functional parts for the object, and SAM3 segments the described regions. 
Part masks are converted into bounding boxes and attached to their parent object, producing object-conditioned part annotations.

\paragraph{Stage 3: Affordance parsing.}
Affordance supervision is constructed on top of object-centric regions. 
GPT-5 generates action-oriented region descriptions, such as handles, rims, lids, or buttons, together with the affordance labels. 
SAM3 segments these described regions, after which we filter low-confidence masks, convert valid masks into affordance boxes, and sample interaction points. 

\paragraph{Hierarchy reconstruction and quality control.}
Since object, part, and affordance annotations are produced by separate stages, we reconstruct them into a strict object-part-affordance hierarchy. 
Affordances are assigned to parts using textual matching when the interaction part is explicitly specified, and geometric matching otherwise, based on point containment and region overlap. 
We further apply consistency checks to ensure valid parent-child relations, geometric containment between levels, and removal of highly overlapping duplicates. 

\subsection{Benchmark Statistics and Diversity}
\label{sec:dataset_statistics}

\input{Figures/benchmark}

We construct \textbf{SceneParser-Bench} based on EgoObject~\cite{zhu2023egoobjects} as a large-scale benchmark for hierarchical scene parsing. 
It contains 110K training and 5K validation images with no image overlap, annotated with explicit hierarchies linking localized objects, semantic parts, and affordances. 
The train/val splits include 743K/33.9K objects, 1.1M/49.5K parts, and 1.67M/75.9K affordances, totaling 1.74M valid object-part-affordance chain instances.

\autoref{fig:benchmark_overview} summarizes the benchmark statistics. 
The three word clouds show broad vocabularies at the object, part, and affordance/action levels. 
The right panel summarizes split scale, category counts, and hierarchy density, showing stable structural density across train and validation splits. 
The top-30 object-part-affordance composition frequencies further reveal diverse long-tailed combinations beyond frequent head patterns.

\input{Figures/model}
\section{SceneParser}
\label{sec:sceneparser}

\label{sec:model_overview}

SceneParser is a VLM-based hierarchical parser built on Rex-Omni~\citep{jiang2025detect}. 
Given an RGB image, it autoregressively generates a unified object-part-affordance hierarchy following~\autoref{sec:Problem_Formulation}. 
Unlike perception-stitching pipelines that predict objects, parts, and affordances independently, SceneParser directly decodes a structured hierarchy with explicit parent-child bindings. 
Spatial outputs reuse Rex-Omni's 1000-bin relative coordinate tokens, with serialization details in~\autoref{appendix:serialization}. 
As shown in~\autoref{fig:model_overview}, we adapt this backbone by training it on serialized JSON-style hierarchy targets, together with structural-completion pseudo labels and curriculum learning.

\subsection{Structural Completion with Pseudo Labels}
\label{sec:structural_completion}

Real scenes contain mixed-completeness supervision: some objects have reliable parts and affordances, while others are annotated only at higher levels. 
Training on such partial hierarchies can produce broken parent-child structures or uncontrolled over-expansion. 
We therefore introduce structural-completion pseudo labels that convert each target into a valid object-part-affordance tree.

For an object without valid parts, we insert a placeholder part label \texttt{\_\_placeholder\_part\_\_} with an object-inherited box. 
For a part without valid affordances, we insert a placeholder affordance label \texttt{\_\_placeholder\_action\_\_} with a fallback spatial field. 
These placeholders enforce parent-child consistency without serving as fine-grained semantic ground truth. 
They encode valid non-expandable cases, regularize the serialized output space, and reduce invalid hierarchical generations. During evaluation, placeholder entries are removed and are not counted as semantic parts or affordances.

\subsection{Curriculum Learning}
\label{sec:pseudo_curriculum}

Structural-completion pseudo labels improve hierarchical completeness, but using them throughout training may dilute fine-grained grounding signals, as placeholders are not semantic ground truth. 
Inspired by curriculum learning~\citep{bengio2009curriculum,kumar2010self,platanios2019competence} and curriculum-style multimodal training~\citep{li2023llava,wu2024curriculum}, we use a three-stage curriculum: first learn reliable localization from non-pseudo labels, then gradually introduce pseudo-completed targets for structural completion.

Stage 1 uses only non-pseudo supervision to stabilize object localization, part grounding, and affordance pointing. 
Stage 2 mixes 70\% non-pseudo and 30\% pseudo-completed samples to encourage structural expansion while preserving grounding quality. 
Stage 3 increases the pseudo ratio to 50\% to reinforce hierarchy completion and parse consistency. 
Training continues across stages, with each stage initialized from the previous one and trained with stage-wise learning-rate decay.



%% file: Figures/benchmark.tex
\begin{figure*}[t]
    \centering
    \includegraphics[width=\textwidth]{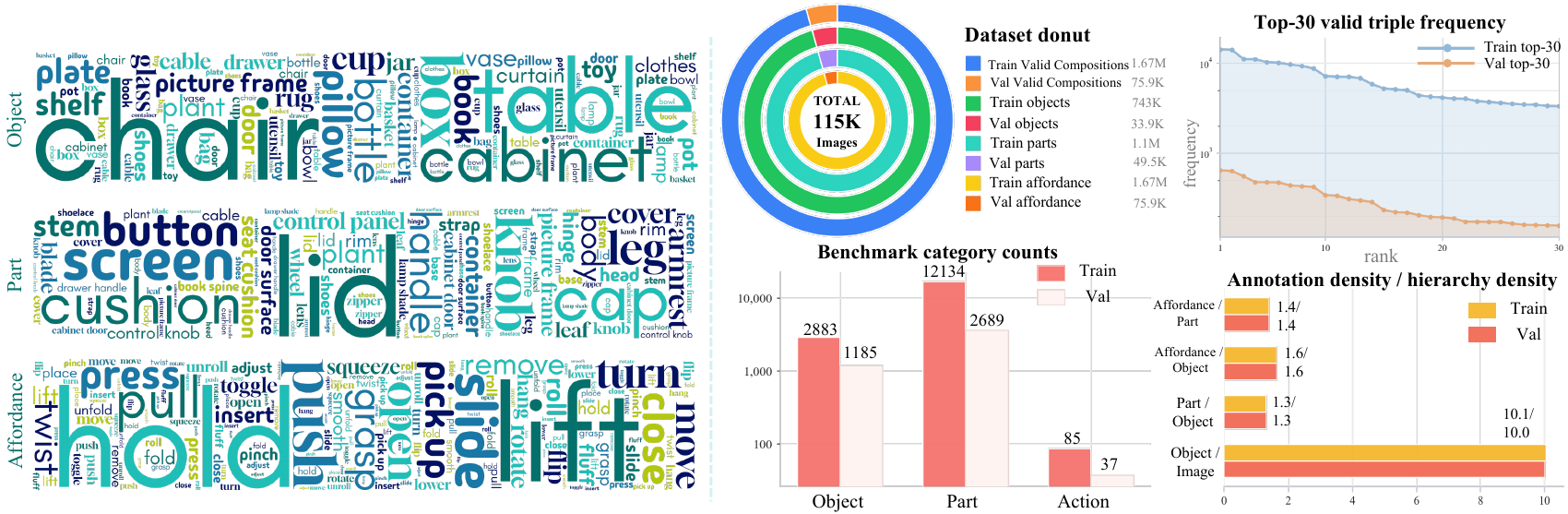}
    \caption{
    Statistical overview of SceneParser-Bench.
    Left: object, part, and affordance/action word clouds from top to bottom.
    Right: split scale, category counts, hierarchy density, and top-30 object-part-affordance compositions.
    The benchmark provides broad coverage, stable hierarchy density, and diverse long-tailed compositions.
    }
    \label{fig:benchmark_overview}
\end{figure*}

%% file: Figures/model.tex
\begin{figure*}[t]
    \centering
    \includegraphics[width=\textwidth]{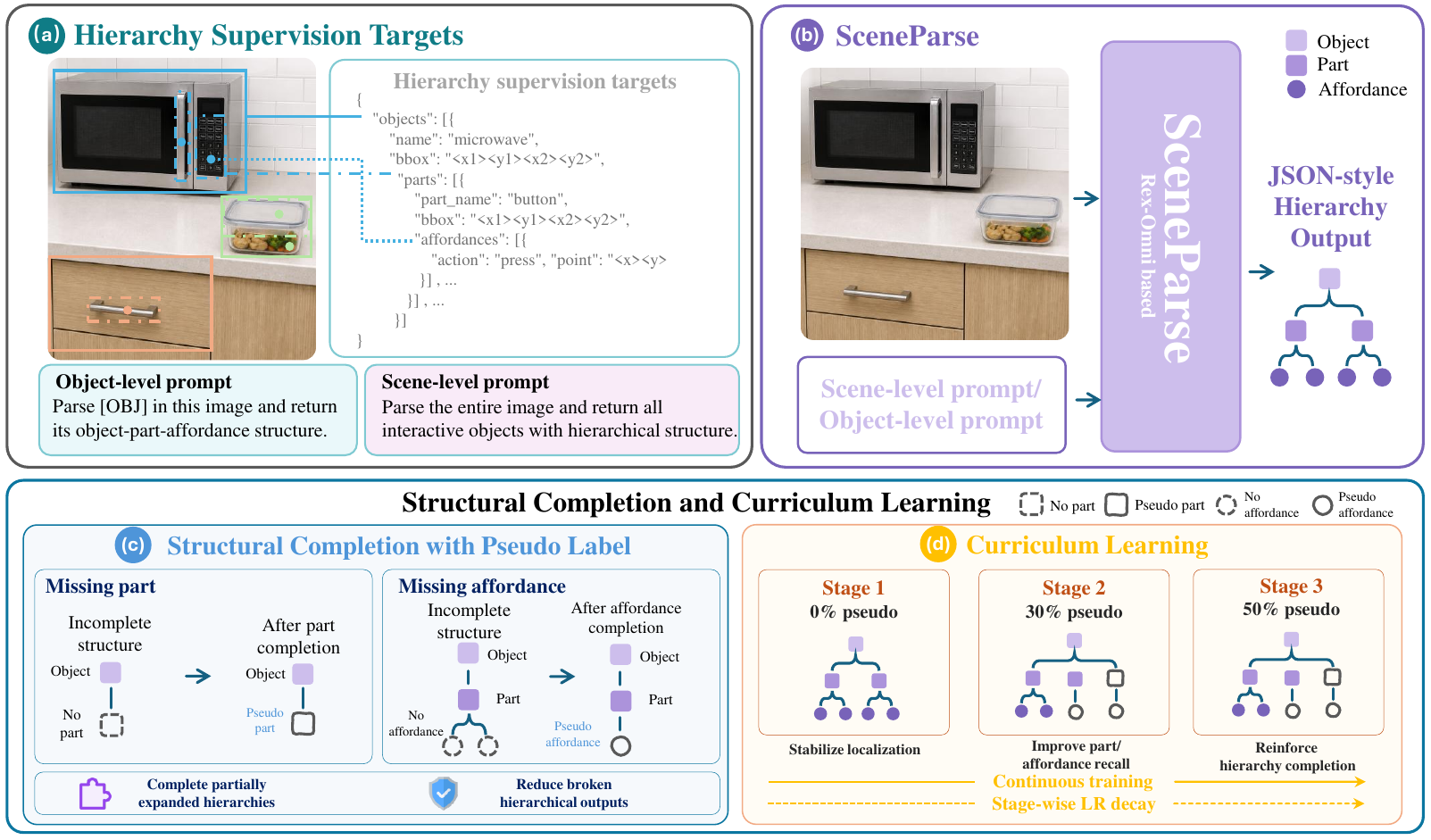}
    \caption{
    Overview of SceneParser.
    SceneParser learns serialized object-part-affordance hierarchy targets and generates JSON-style hierarchies.
    Structural-completion pseudo labels complete partial hierarchies, while curriculum learning balances grounding reliability and hierarchical completeness.
    }
    \vspace{-12pt}
    \label{fig:model_overview}
\end{figure*}

%% file: Sections/4_exp.tex
\input{Tables/0_compare}
\section{Experiments}

We evaluate SceneParser along four axes: 
(1) hierarchy-aware performance on SceneParser-Bench; 
(2) the effect of explicit hierarchy, structural completion, and curriculum learning; 
(3) transfer to conventional object detection and affordance grounding; and 
(4) the utility of parsed hierarchies for downstream interaction-oriented reasoning.

\subsection{Main Results on SceneParser-Bench}
\label{sec:exp_main_benchmark}

We evaluate two settings: \textit{object-level} parsing, where the model receives an image and target object category, and \textit{scene-level} parsing, where it parses all visible interactive objects from the image alone. 
Both require structured hierarchy generation following the schema in~\autoref{appendix:prompts}. 
We compare closed-source MLLMs, open-source MLLMs, and a Rex-Omni stitching baseline, where objects, parts, and affordances are predicted separately and then merged into object-part-affordance hierarchies. 
We report L1-L3 and ParseRate defined in~\autoref{sec:evaluation_suite} with training details in~\autoref{appendix:training_impl}.

\paragraph{Results.}
As shown in~\autoref{tab:hierarchical_compare_compact}, existing MLLMs obtain non-trivial L1 scores but drop sharply at L2 and L3, showing that object grounding alone does not yield valid object-part-affordance chains. 
ParseRate further shows that structural completion is also difficult: although some MLLMs expand many objects, their low L2/L3 scores indicate incomplete or incorrectly bound part-affordance structures. 
Rex-Omni Stitching achieves high object-level L1 precision, but its low ParseRate and missing L3 chains expose the limitation of stitching independent predictors. 
SceneParser achieves the strongest L2/L3 performance and the highest ParseRate in both object-level and scene-level settings, demonstrating better cross-level binding and structural completeness.

\subsection{Ablation Studies}
\label{sec:exp_ablation}

We ablate the key representation and training choices of SceneParser, including explicit hierarchical generation, structural-completion pseudo labels, and curriculum learning. 
We also analyze whether object and part context improves affordance grounding.

\subsubsection{Effect of Hierarchical Generation and Affordance Context}
\label{sec:exp_hierarchy_affordance}

\paragraph{Setup.}
We first isolate the effect of hierarchical output organization by comparing two variants with identical prediction fields: \textit{Flat Triplets}, which predicts object, part, and affordance fields as independent triplets, and \textit{Nested Hierarchy}, which directly preserves object-part-affordance parent-child bindings. 
For fair evaluation, flat triplets are deterministically converted into object-part-affordance hierarchies, and both variants are evaluated with the same structure-aware L1-L3 protocol as the main benchmark; details are provided in~\autoref{appendix:flat_to_hierarchy}.

We then test whether affordance grounding benefits from structural context by comparing \textit{Direct Point}, \textit{Object + Point}, and \textit{Object + Part + Point}. 
All variants are evaluated with the same point-validity criterion as the main evaluation, and we report affordance precision, recall, and F1.

\input{Figures/hierarchy_ablation}
\input{Tables/curriculum_ablation}

\paragraph{Results.}
As shown in~\autoref{fig:hierarchy_ablation}(a), \textit{Nested Hierarchy} outperforms \textit{Flat Triplets} under identical flattened evaluation, improving L1 from 40.7 to 73.5, L2 from 22.3 to 42.9, and L3 from 17.8 to 29.3.
Since both variants predict identical fields, the gain comes from parent-child organization rather than additional information, showing that hierarchy helps generate coherent object-part-affordance chains.

\autoref{fig:hierarchy_ablation}(b) shows that affordance grounding improves as more structural context is provided. 
Object context improves affordance F1 from 40.6 to 41.7, while adding part context achieves the best F1 of 42.8 and raises recall from 34.7 to 41.9. 
This suggests that affordances are better grounded within explicit object-part structures than as isolated points.

\subsubsection{Effect of Structural-Completion Curriculum}
\label{sec:exp_curriculum}

\paragraph{Setup.}
We evaluate the trade-off between reliable grounding and structural completion. 
All variants use the same architecture and evaluation protocol. 
\textit{Ours-w/o pseudo} uses only non-pseudo supervision, \textit{Ours-pseudo} uses pseudo-completed supervision throughout, and \textit{Ours-CL} applies our three-stage curriculum that progressively mixes non-pseudo and pseudo-completed samples. For fair comparison, placeholder entries are removed before computing L1-L3 and ParseRate.

\paragraph{Results.}
As shown in~\autoref{tab:curriculum_ablation}, \textit{Ours-CL} achieves the best ParseRate, improving structural completeness by +2.7 points over the best single-regime baseline. 
It also obtains the best L3 score, indicating stronger affordance-level chain recovery. 
Although L1 and L2 slightly decrease, the drops are small, suggesting that curriculum learning improves hierarchical completeness without substantially weakening object and part grounding. 
These results show that non-pseudo supervision stabilizes localization, pseudo-completed targets encourage valid expansion, and curriculum learning balances the two.

\subsection{Cross-Task Transfer}
\label{sec:exp_transfer}

We evaluate whether SceneParser preserves conventional grounding ability beyond SceneParser-Bench on COCO object detection~\citep{lin2014microsoft} and AGD20K affordance grounding~\citep{luo2022learning}.

\paragraph{COCO object detection.}
We extract object boxes from SceneParser's structured outputs and evaluate them under the COCO detection protocol. 
As shown in~\autoref{tab:coco_detection_with_ours}, SceneParser achieves 66.8 F1@0.5 from structured outputs, remaining competitive with MLLM-based detectors despite not being optimized as a standalone detector. 
This suggests that object-part-affordance training preserves transferable object localization.

\input{Tables/COCO_Compare}

\paragraph{AGD20K affordance grounding.}
We evaluate affordance point prediction on AGD20K using point-in-mask accuracy over samples with valid ground-truth masks. 
As shown in~\autoref{fig:hierarchy_ablation}(c), SceneParser outperforms Affordance-R1 on both seen objects (87.67 vs. 60.79) and unseen objects (82.79 vs. 57.53). 
These results show that SceneParser transfers to conventional detection and affordance grounding while retaining richer object-part-affordance outputs.

\vspace{-7pt}
\subsection{Downstream Interaction-Oriented Reasoning}
\label{sec:exp_downstream_reasoning}

We further test whether parsed hierarchies can serve as decision-ready inputs for downstream planning. 
Given the same image and task instruction, we compare a task-only prompt, which may already specify target object semantics, with a structure-augmented prompt that additionally provides the SceneParser hierarchy. 
As shown in~\autoref{appendix:planning_probe}, the hierarchy adds explicit object-part-affordance bindings, helping the planner produce more consistent interaction targets and more coherent manipulation steps. 
This qualitative probe suggests that SceneParser provides an actionable intermediate representation beyond task-level object semantics and local perception.

%% file: Tables/0_compare.tex
\begin{table*}[t]
  \centering
  \caption{
    Hierarchical parsing results across models.
    Entries are object-level/scene-level percentages.
    Bold and underlined values denote the best and second-best results per metric and setting.
    }
  \label{tab:hierarchical_compare_compact}
  \scriptsize
  \setlength{\tabcolsep}{2.2pt}
  \renewcommand{\arraystretch}{1.03}
  \providecommand{\os}[2]{\mbox{#1/#2}}
  \providecommand{\best}[1]{\textbf{#1}}
  \providecommand{\second}[1]{\underline{#1}}

  \resizebox{\textwidth}{!}{%
  \begin{tabular}{@{}l ccc ccc ccc c@{}}
    \toprule
    \multirow{2}{*}{\textbf{Method}}
    & \multicolumn{3}{c}{\textbf{L1}}
    & \multicolumn{3}{c}{\textbf{L2}}
    & \multicolumn{3}{c}{\textbf{L3}}
    & \multirow{2}{*}{\textbf{\shortstack{Parse\\Rate}}} \\
    \cmidrule(lr){2-4}
    \cmidrule(lr){5-7}
    \cmidrule(lr){8-10}
    & \textbf{P} & \textbf{R} & \textbf{F1}
    & \textbf{P} & \textbf{R} & \textbf{F1}
    & \textbf{P} & \textbf{R} & \textbf{F1}
    &  \\
    \midrule

    \multicolumn{11}{@{}l}{\textit{Closed-source MLLMs}} \\
    GPT-5.5~\citep{openai2025gpt5}
    & \os{23.8}{17.0} & \os{19.5}{14.1} & \os{21.5}{15.4}
    & \os{1.3}{2.1} & \os{3.0}{3.5} & \os{1.8}{2.6}
    & \os{0.0}{0.7} & \os{0.0}{1.3} & \os{0.0}{1.3}
    & \os{25.3}{18.7} \\

    Gemini 3.1 Pro~\citep{google2026gemini31pro}
    & \os{67.3}{\second{47.9}} & \os{48.4}{12.8} & \os{\second{56.3}}{20.2}
    & \os{11.0}{\second{11.9}} & \os{6.4}{4.8} & \os{8.1}{\second{6.8}}
    & \os{\second{0.2}}{\second{5.6}} & \os{\second{0.1}}{\second{2.1}} & \os{\second{0.1}}{\second{3.0}}
    & \os{28.6}{16.2} \\

    Qwen3.5-plus~\citep{qwen2026qwen35}
    & \os{25.9}{33.5} & \os{\second{52.2}}{\second{19.8}} & \os{34.6}{\second{24.9}}
    & \os{3.1}{6.4} & \os{10.0}{\second{5.2}} & \os{4.7}{5.7}
    & \os{0.0}{2.4} & \os{\second{0.1}}{2.0} & \os{\second{0.1}}{2.2}
    & \os{\second{64.4}}{\second{24.5}} \\

    \midrule
    \multicolumn{11}{@{}l}{\textit{Open-source MLLMs}} \\
    Qwen2.5-VL-72B~\citep{bai2025qwen25vl}
    & \os{55.2}{37.6} & \os{34.1}{12.8} & \os{42.1}{19.1}
    & \os{4.9}{4.9} & \os{2.4}{1.6} & \os{3.2}{2.4}
    & \os{0.1}{1.4} & \os{0.0}{0.4} & \os{\second{0.1}}{0.6}
    & \os{29.5}{13.9} \\

    Qwen3-VL-235B~\citep{qwen2025qwen3vl}
    & \os{21.2}{3.6} & \os{41.8}{2.3} & \os{28.1}{2.8}
    & \os{3.1}{0.3} & \os{8.7}{0.2} & \os{4.6}{0.2}
    & \os{0.0}{0.0} & \os{0.0}{0.0} & \os{0.0}{0.0}
    & \os{49.6}{2.4} \\

    Qwen3.5-35B~\citep{qwen2026qwen35}
    & \os{32.9}{19.4} & \os{48.8}{11.2} & \os{39.3}{14.2}
    & \os{3.8}{2.9} & \os{9.6}{2.3} & \os{5.4}{2.6}
    & \os{0.0}{0.6} & \os{\second{0.1}}{0.5} & \os{0.0}{0.5}
    & \os{60.8}{13.5} \\

    SEED2.0~\citep{seed202620}
    & \os{41.6}{10.4} & \os{34.8}{6.6} & \os{37.9}{8.1}
    & \os{5.7}{2.3} & \os{4.4}{0.7} & \os{5.0}{1.1}
    & \os{0.0}{0.3} & \os{0.0}{0.1} & \os{0.0}{0.2}
    & \os{26.0}{4.0} \\

    Qwen3.5-4B~\citep{qwen2026qwen35}
    & \os{2.5}{0.0} & \os{0.0}{0.0} & \os{0.0}{0.0}
    & \os{0.0}{0.0} & \os{0.0}{0.0} & \os{0.0}{0.0}
    & \os{0.0}{0.0} & \os{0.0}{0.0} & \os{0.0}{0.0}
    & \os{0.0}{0.0} \\

    \midrule
    \multicolumn{11}{@{}l}{\textit{Rex-Omni based methods}} \\
    Rex-Omni Stitching~\citep{jiang2025detect}
    & \os{\best{78.3}}{$-$} & \os{24.2}{$-$} & \os{37.0}{$-$}
    & \os{\best{76.1}}{$-$} & \os{\second{13.1}}{$-$} & \os{\second{22.4}}{$-$}
    & \os{0.0}{$-$} & \os{0.0}{$-$} & \os{0.0}{$-$}
    & \os{15.0}{$-$} \\

    \rowcolor[HTML]{D9E1F2}\textbf{Ours}
    & \os{\second{76.8}}{\best{59.1}} & \os{\best{70.4}}{\best{50.7}} & \os{\best{73.5}}{\best{54.6}}
    & \os{\second{43.8}}{\best{38.1}} & \os{\best{42.1}}{\best{36.9}} & \os{\best{42.9}}{\best{37.5}}
    & \os{\best{30.0}}{\best{26.2}} & \os{\best{28.7}}{\best{26.4}} & \os{\best{29.3}}{\best{26.3}}
    & \os{\best{66.3}}{\best{53.2}} \\

    \bottomrule
  \end{tabular}%
  }

\end{table*}

%% file: Figures/hierarchy_ablation.tex
\begin{figure*}[t]
    \centering
    \includegraphics[width=0.95\textwidth]{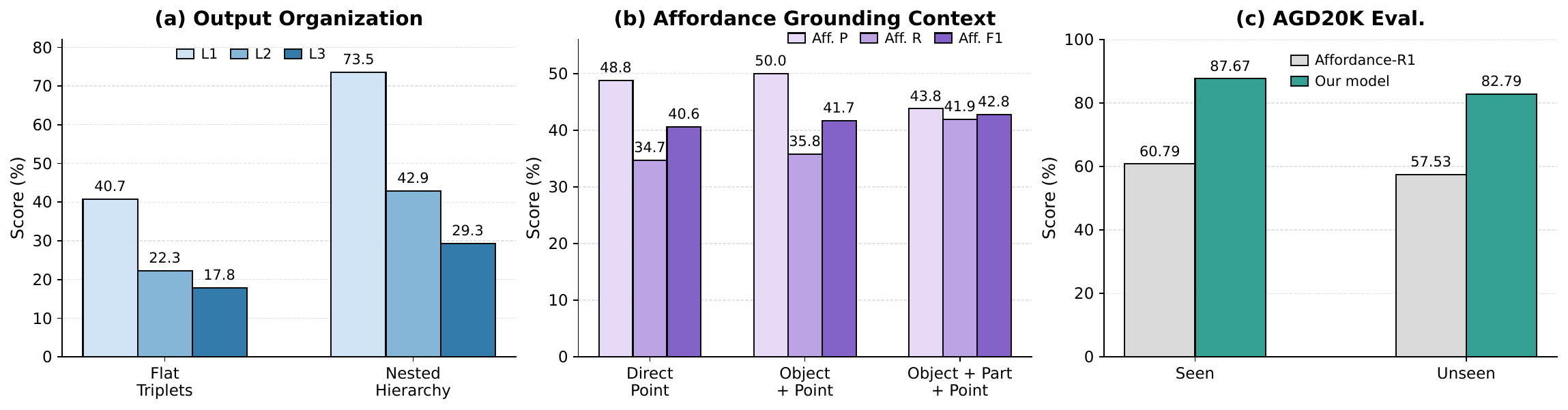}
    \caption{
    Analysis of hierarchy, affordance context, and cross-benchmark transfer.
    (a) Nested hierarchy outperforms flat triplets under identical structure-aware L1-L3 evaluation.
    (b) Adding object and part context improves affordance grounding, especially recall and F1.
    (c) On AGD20K~\citep{luo2022learning}, SceneParser achieves higher point-in-mask accuracy than Affordance-R1~\citep{wang2026affordance} on both seen and unseen object splits.
    Values are percentages.
    }
    \vspace{-8pt}
    \label{fig:hierarchy_ablation}
\end{figure*}

%% file: Tables/curriculum_ablation.tex
\begin{table}[t]
  \centering
  \caption{
    Effect of structural-completion pseudo labels and curriculum learning.
    Values are object-level hierarchical parsing scores in percentages; $\Delta$ is computed against the best single-regime baseline.
    }
  \label{tab:curriculum_ablation}
  \small
  \setlength{\tabcolsep}{13pt}
  \renewcommand{\arraystretch}{1.08}
  \begin{tabular}{lcccc}
    \toprule
    \textbf{Method} & \textbf{L1} & \textbf{L2} & \textbf{L3} & \textbf{ParseRate} \\
    \midrule
    Ours-w/o pseudo & 74.1 & \textbf{43.1} & 29.1 & 39.6 \\
    Ours-pseudo     & \textbf{74.3} & \textbf{43.1} & 28.8 & 40.5 \\
   \rowcolor[HTML]{D9E1F2} Ours-CL         & 73.5 & 42.9 & \textbf{29.3} & \textbf{43.2} \\
    \midrule
    $\Delta$ (CL - best single regime) & -0.8 & -0.2 & +0.2 & +2.7 \\
    \bottomrule
  \end{tabular}
  \vspace{-8pt}
\end{table}

%% file: Tables/COCO_Compare.tex
\begin{table*}[t]
  \centering
  \caption{
    COCO object detection transfer results.
    Object boxes from SceneParser are extracted from structured JSON outputs and evaluated with the standard COCO protocol~\citep{lin2014microsoft}.
    Baselines cover closed-set detectors, open-set detectors, and MLLM-based detectors.
    }
  \label{tab:coco_detection_with_ours}
  \scriptsize
  \setlength{\tabcolsep}{4pt}
  \renewcommand{\arraystretch}{0.92}
  \resizebox{\textwidth}{!}{
  \begin{tabular}{llrrrrrr}
    \toprule
    Type & Method
    & R@0.5 & P@0.5 & F1@0.5 
    & R@.95 & P@.95 & F1@.95 \\
    \midrule
    Closed-set & DINO-R50~\citep{zhang2022dino} 
    & 62.6 & 76.5 & 68.8 & 17.8 & 25.8 & 21.1 \\
    Closed-set & DINO-Swin-L~\citep{zhang2022dino}  
    & 69.8 & 82.5 & 75.6 & 22.2 & 29.7 & 25.4 \\
    Open-set   & Grounding DINO-Swin-T~\citep{liu2024grounding} 
    & 62.8 & 78.4 & 69.8 & 20.5 & 26.2 & 23.0 \\
    \midrule
    MLLM       & Qwen2.5-VL-7B~\citep{bai2025qwen25vl}
    & 75.8 & 55.4 & 64.0 & 10.6 & 15.5 & 12.6 \\
    MLLM       & Qwen2.5-VL-3B~\citep{bai2025qwen25vl}
    & 77.2 & 55.7 & 64.7 & 12.7 & 18.3 & 15.0 \\
    MLLM       & SEED1.5-VL~\citep{guo2025seed15vl}
    & 78.6 & 65.3 & 71.3 & 12.7 & 16.4 & 14.3 \\
    MLLM       & Rex-Omni-SFT~\citep{jiang2025detect}
    & 70.1 & 66.4 & 68.2 & 14.8 & 17.0 & 15.8 \\
    \rowcolor[HTML]{D9E1F2}
    MLLM       & \textbf{Ours}
    & 65.3 & 68.3 & 66.8 & 13.4 & 12.8 & 13.1 \\
    \bottomrule
  \end{tabular}
  }
  \vspace{-12pt}
\end{table*}

%% file: Sections/5_con.tex
\vspace{-7pt}
\section{Conclusion and Limitations}
\label{sec:conclusion}

\vspace{-7pt}
We introduced \textit{Hierarchical Scene Parsing}, an interaction-oriented scene understanding task that represents physical scenes as explicit \textit{scene $\rightarrow$ object $\rightarrow$ part $\rightarrow$ affordance} hierarchies. 
Unlike conventional localization outputs, this formulation requires cross-level bindings, grounding each affordance in the functional part and object that support it. 
We instantiated this task with SceneParser-Bench, a large-scale hierarchical benchmark, and SceneParser, a VLM-based parser trained with unified hierarchical generation, structural-completion pseudo labels, and curriculum learning. 
Our structure-aware evaluation shows that existing MLLMs and perception-stitching pipelines struggle to recover valid object-part-affordance chains, while SceneParser achieves stronger hierarchical parsing, transfers to conventional detection and affordance grounding, and provides useful structure for downstream interaction-oriented reasoning.

\paragraph{Limitations.}
SceneParser is a first step toward decision-ready hierarchical scene understanding. 
SceneParser-Bench may inherit noise from its automatic data engine, despite filtering and quality control. 
The current representation is image-centric, using 2D boxes and points without modeling 3D geometry, dynamics, force constraints, or manipulation trajectories. 
Our planning probe is qualitative, and future work should evaluate closed-loop robotic execution. 
Long-tailed object-part-affordance compositions also require stronger compositional generalization.

%% file: Sections/X_app.tex
\appendix

\section{Formal Definition of Evaluation Metrics}
\label{appendix:evaluation_details}

\paragraph{Overview.}
We provide the formal definition of the structure-aware evaluation protocol used in~\autoref{sec:evaluation_suite}. 
The protocol evaluates three conditional levels: L1 for objects, L2 for object-part chains, and L3 for full object-part-affordance chains. 
At all levels, predictions are matched to ground truth using greedy one-to-one assignment, and precision, recall, and F1 are computed from accumulated true positives, false positives, and false negatives.

\paragraph{Precision, recall, and F1.}
For each level $\ell \in \{1,2,3\}$ and IoU threshold $\tau$, we compute
\[
\mathrm{P}_{\ell,\tau}
=
\frac{\mathrm{TP}_{\ell,\tau}}
{\mathrm{TP}_{\ell,\tau}+\mathrm{FP}_{\ell,\tau}},
\quad
\mathrm{R}_{\ell,\tau}
=
\frac{\mathrm{TP}_{\ell,\tau}}
{\mathrm{TP}_{\ell,\tau}+\mathrm{FN}_{\ell,\tau}},
\]
\[
\mathrm{F1}_{\ell,\tau}
=
\frac{
2\mathrm{P}_{\ell,\tau}\mathrm{R}_{\ell,\tau}
}{
\mathrm{P}_{\ell,\tau}+\mathrm{R}_{\ell,\tau}
}.
\]
Here $\mathrm{TP}_{\ell,\tau}$, $\mathrm{FP}_{\ell,\tau}$, and $\mathrm{FN}_{\ell,\tau}$ are accumulated over the evaluation set under the hierarchical matching constraints defined below.

\paragraph{L1: object matching.}
At L1, predicted objects are matched to ground-truth objects. 
A predicted object is counted as a true positive if its object name matches the ground-truth object name and its object box satisfies
\[
\mathrm{IoU}(b_o^{pred}, b_o^{gt}) \ge \tau .
\]
Unmatched predicted objects are counted as false positives, and unmatched ground-truth objects are counted as false negatives.

\paragraph{L2: object-part matching.}
At L2, part matching is performed only within correctly matched object pairs from L1. 
A predicted object-part chain is counted as a true positive if the parent object is correctly matched, the part name matches, and the part box satisfies
\[
\mathrm{IoU}(b_p^{pred}, b_p^{gt}) \ge \tau .
\]
Thus, a part can be correct only when its parent object is also correct. 
Unmatched predicted parts under valid matched objects are counted as false positives, and unmatched ground-truth parts under matched objects are counted as false negatives.

\paragraph{L3: object-part-affordance matching.}
At L3, affordance matching is performed only within valid object-part matches from L2. 
A predicted object-part-affordance chain is counted as a true positive if the parent object and part are correctly matched, the action label matches, and the predicted interaction point is spatially valid:
\[
u^{pred} \in R_a^{gt},
\]
where $R_a^{gt}$ denotes the ground-truth valid region, defined as the ground-truth affordance box when available and otherwise the parent part box.
Thus, L3 correctness requires object localization, part localization, action recognition, point grounding, and valid cross-level binding.

\paragraph{ParseRate.}
ParseRate measures whether a matched object is structurally completed with its required lower-level hierarchy. 
We define a ground-truth object as \emph{parse-eligible} if it contains parts and/or affordances and therefore requires structural expansion. 
Given a parse-eligible GT object that is correctly matched at IoU threshold $0.5$, its prediction is counted as complete if it satisfies the required expansion pattern: it predicts parts when GT parts exist, affordances when GT affordances exist, and both when both are required.

Let $\mathcal{E}$ denote the set of parse-eligible GT objects in the evaluation set, and let $\mathcal{C}\subseteq\mathcal{E}$ denote the subset whose matched predictions satisfy the required structural expansion. 
We define
\[
\mathrm{ParseRate}_{0.5}
=
\frac{|\mathcal{C}|}{|\mathcal{E}|}.
\]

We report ParseRate under two settings. 
For object-level evaluation, $\mathcal{E}$ contains parse-eligible target objects specified by the input category. 
For scene-level evaluation, $\mathcal{E}$ contains all parse-eligible objects across all scenes. 
Thus, object-level ParseRate measures target-conditioned hierarchy completion, while scene-level ParseRate measures full-scene hierarchical coverage and consistency.

\section{Data Engine Illustration}
\label{appendix:data_engine_figure}

\begin{center}
    \includegraphics[width=\textwidth]{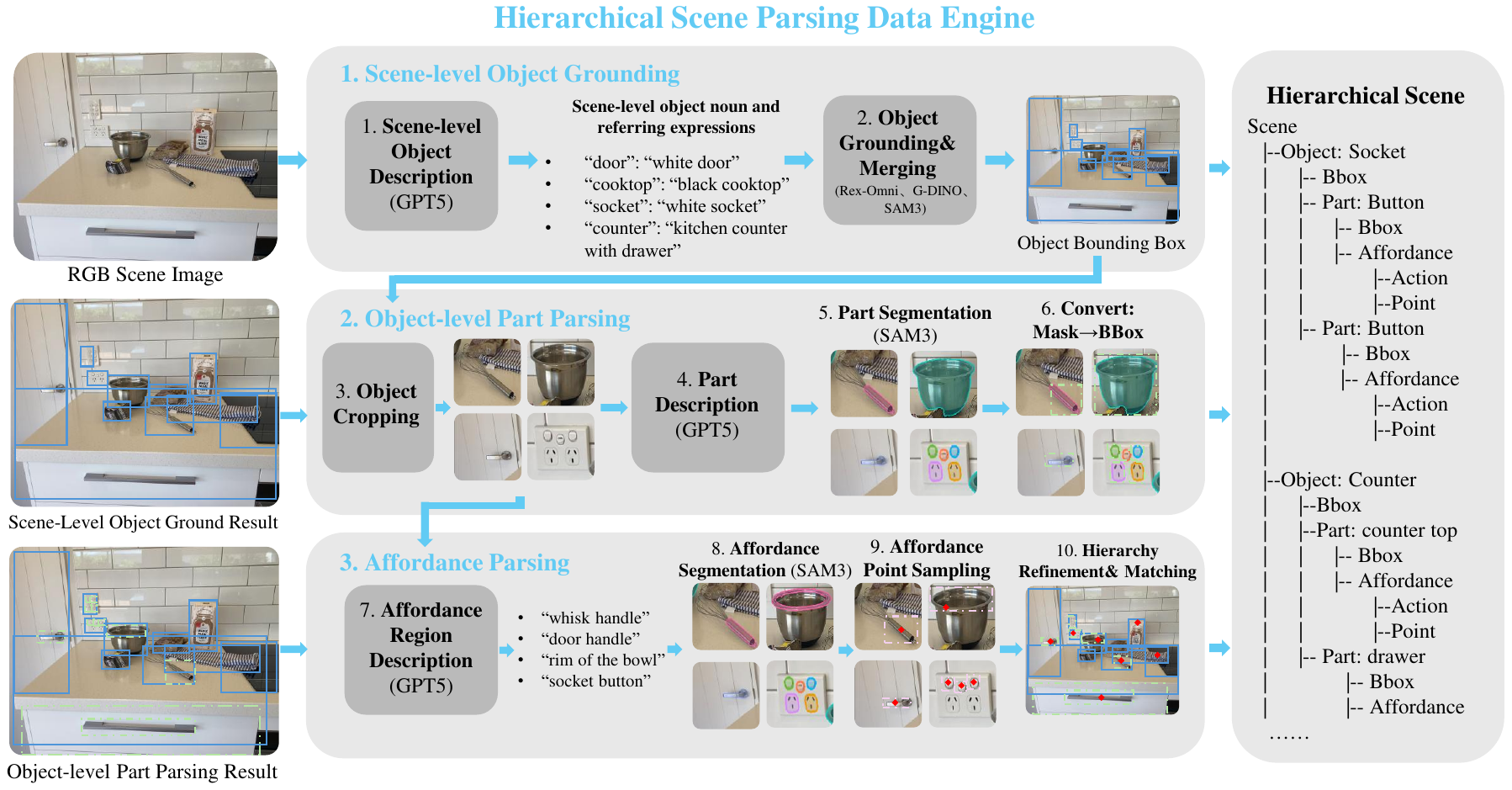}
    \captionof{figure}{
    Overview of the proposed data engine. Starting from an RGB scene image, the engine progressively constructs hierarchical supervision through object-level grounding, object-level part parsing, affordance parsing, and hierarchy reconstruction. The final output is organized as a unified \textit{scene} $\rightarrow$ \textit{object} $\rightarrow$ \textit{part} $\rightarrow$ \textit{affordance} structure for training and evaluation.
    }
    \label{fig:data_engine}
\end{center}

\section{Output Serialization and Coordinate Tokenization}
\label{appendix:serialization}

SceneParser generates a serialized JSON-style hierarchy following the object-part-affordance formulation in~\autoref{sec:Problem_Formulation}. 
A compact training target is represented as:

\begin{promptbox}{Serialized Object-Part-Affordance Target}
{
  "objects": [{
    "name": "microwave",
    "bbox": "<x1><y1><x2><y2>",
    "parts": [{
      "part_name": "door",
      "bbox": "<x1><y1><x2><y2>",
      "affordances": [
        {"action": "open",  "point": "<x><y>"},
      ]
    }, {
      "part_name": "button panel",
      "bbox": "<x1><y1><x2><y2>",
      "affordances": [
        {"action": "press", "point": "<x><y>"}
      ]
    }]
  }, {
    "name": "drawer",
    "bbox": "<x1><y1><x2><y2>",
    "parts": [{
      "part_name": "handle",
      "bbox": "<x1><y1><x2><y2>",
      "affordances": [
        {"action": "pull", "point": "<x><y>"}
      ]
    }]
  }]
}
\end{promptbox}

Spatial outputs follow the Rex-Omni coordinate-token protocol~\citep{jiang2025detect}. 
For each image, continuous coordinates are first converted to relative image coordinates and quantized into 1000 discrete bins, corresponding to positions from 0 to 999 along each image axis. 
Each bin is represented by a dedicated coordinate token, so a bounding box is serialized as four spatial tokens $(x_1,y_1,x_2,y_2)$ and an affordance point as two spatial tokens $(x,y)$. 
These coordinate tokens are inserted directly into the JSON-style hierarchy and generated together with object names, part names, and affordance actions in a single autoregressive sequence. 
This allows SceneParser to decode a unified hierarchy rather than separate object, part, and affordance predictions.

\section{Training and Implementation Details}
\label{appendix:training_impl}

We use the serialization and coordinate-token protocol described in~\autoref{appendix:serialization}. 
This section provides the exact curriculum schedule and training hyperparameters used for SceneParser.

All stages use AdamW with cosine learning-rate decay, warmup ratio 0.03, weight decay 0.01, maximum gradient norm 1.0, bf16 mixed precision, gradient checkpointing, and maximum sequence length 4096. 
We train with per-device batch size 2 and gradient accumulation 8 on 16 A100-80GB GPUs across two nodes, giving an effective global batch size of 256. 
The complete three-stage curriculum takes about 24 wall-clock hours on this setup, corresponding to roughly 384 GPU-hours.
Training is implemented with \texttt{torchrun} and DeepSpeed ZeRO-2.
We provide the full training scripts, DeepSpeed configurations, and stage-wise launch commands in the released codebase to facilitate reproducibility.

\begin{table}[h]
\centering
\caption{
Training schedule of the structural-completion curriculum.
Stages are trained sequentially, with each stage initialized from the previous one.
}
\label{tab:training_schedule}
\small
\setlength{\tabcolsep}{4pt}
\renewcommand{\arraystretch}{1.05}
\begin{tabular}{lcccc}
\toprule
Stage & Supervision & Epochs & Main/MM LR & Vision LR \\
\midrule
Stage 1 & No-pseudo & 3 & $2\times10^{-5}$ & $2\times10^{-6}$ \\
Stage 2 & 70\% non-pseudo / 30\% pseudo & 4 & $1\times10^{-5}$ & $1\times10^{-6}$ \\
Stage 3 & 50\% non-pseudo / 50\% pseudo & 3 & $6\times10^{-6}$ & $6\times10^{-7}$ \\
\bottomrule
\end{tabular}
\end{table}

\section{Evaluation Prompts}
\label{appendix:prompts}

\subsection{Object-Level Prompt}

\begin{promptbox}{Object-Level Prompt}
You are given one image.

Target category: [OBJ]

Return JSON only with schema:
{
  "objects": [
    {
      "name": "string",
      "bbox": [x1,y1,x2,y2],
      "parts": [
        {
          "part_name": "string",
          "bbox": [x1,y1,x2,y2],
          "affordances": [
            {"action":"string","point":[x,y]}
          ]
        }
      ]
    }
  ]
}

Rules:
- Focus on [OBJ] only.
- If absent, return {"objects": []}.
- No extra keys. No explanation.
\end{promptbox}

\subsection{Scene-Level Prompt}

\begin{promptbox}{Scene-Level Prompt}
You are given one image.

Return JSON only, using exactly this schema:
{
  "objects":[{
    "name":"string","bbox":[x1,y1,x2,y2],
    "parts":[{
      "part_name":"string","bbox":[x1,y1,x2,y2],
      "affordances":[
        {"action":"string","point":[x,y]}
      ]}
    ]}
  ]
}

Constraints:
1. Parse all visible interactive objects.
2. If no valid object, return {"objects":[]}.
3. Always include parts and affordances arrays (empty if none).
4. Do not output extra keys or explanations.
5. No placeholder literals like "string", "object", "part", "action".
\end{promptbox}

\section{Qualitative Visualization}
\label{appendix:qualitative_visualization}

\begin{center}
    \includegraphics[width=\textwidth]{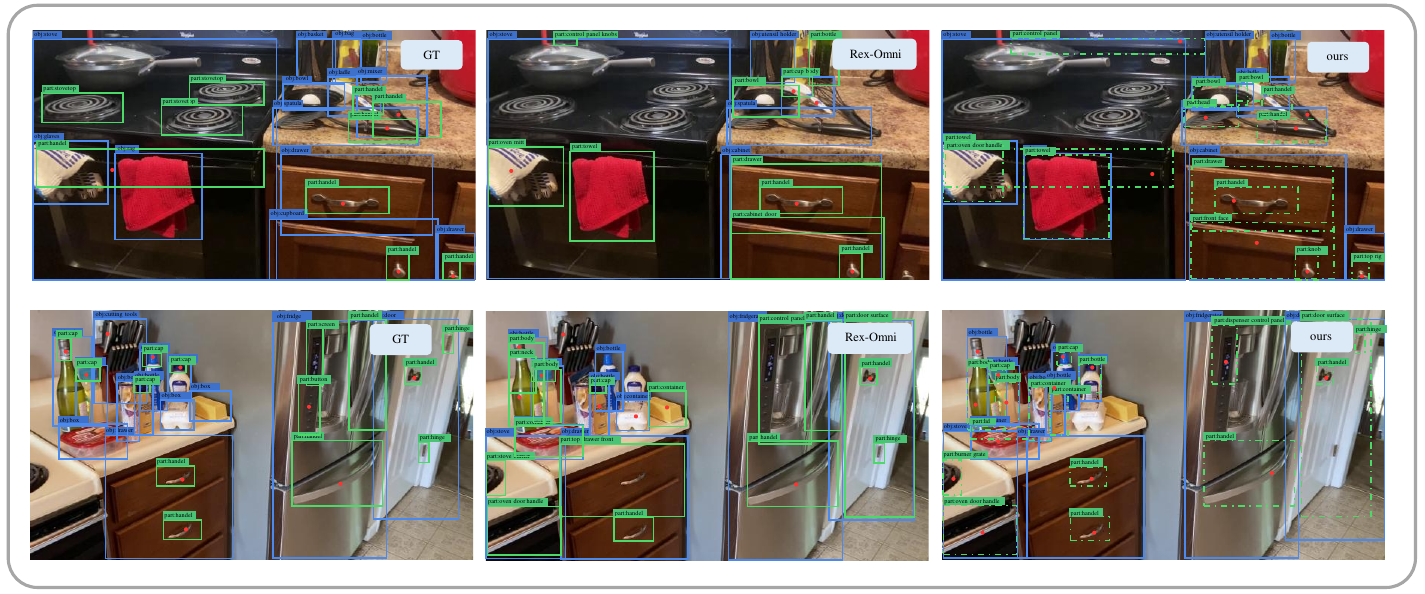}
    \captionof{figure}{
    Qualitative comparison among ground truth, Rex-Omni Stitching, and SceneParser.
    Compared with the stitching baseline, SceneParser produces more coherent object-part-affordance hierarchies with better parent-child bindings and affordance grounding.
    }
    \label{fig:qualitative_visualization}
\end{center}

\section{Flat-Triplet Output and Hierarchy Conversion}
\label{appendix:flat_to_hierarchy}

\paragraph{Flat-triplet output format.}
For the flat-output ablation in~\autoref{sec:exp_ablation}, we remove the explicit parent-child nesting and ask the model to generate a list of object-part-affordance triplets. 
Each triplet contains the object name and box, part name and box, action label, and affordance point.

\paragraph{Why flat triplets?}
The flat-triplet format is designed as a controlled counterpart to the nested hierarchy. 
It preserves the same prediction fields as SceneParser, including object names and boxes, part names and boxes, affordance actions, and interaction points, but removes the explicit object-part-affordance nesting. 
This allows us to isolate the effect of output organization: any performance gap between flat triplets and nested hierarchy mainly reflects whether explicit parent-child structure helps generation, rather than differences in predicted information.

\begin{promptbox}{Flat-Triplet Output Example}
{
  "triplets": [
    {
      "object": "microwave",
      "object_box": "<x1><y1><x2><y2>",
      "part": "door",
      "part_box": "<x1><y1><x2><y2>",
      "action": "open",
      "affordance_point": "<x><y>"
    },
    {
      "object": "microwave",
      "object_box": "<x1><y1><x2><y2>",
      "part": "button panel",
      "part_box": "<x1><y1><x2><y2>",
      "action": "press",
      "affordance_point": "<x><y>"
    },
    {
      "object": "drawer",
      "object_box": "<x1><y1><x2><y2>",
      "part": "handle",
      "part_box": "<x1><y1><x2><y2>",
      "action": "pull",
      "affordance_point": "<x><y>"
    }
  ]
}
\end{promptbox}

\paragraph{Flat-to-hierarchy conversion.}
To compare flat triplets with nested hierarchy under the same structure-aware evaluation, we convert flat predictions into the standard object-part-affordance hierarchy before scoring. 
This deterministic post-processing is applied only to predictions, while the ground truth remains unchanged. 
Each flat triplet contains an object name and box, a part name and box, an action label, and an affordance point. 
We group triplets by \((\text{object name}, \text{object box})\) to create object nodes, then by \((\text{part name}, \text{part box})\) within each object to create part nodes. 
Each \((\text{action}, \text{affordance point})\) pair is attached to its corresponding part, with duplicate affordances removed. 
The converted prediction follows the same \texttt{objects} $\rightarrow$ \texttt{parts} $\rightarrow$ \texttt{affordances} structure as SceneParser outputs.

\noindent\textbf{Evaluation after conversion.}
After conversion, both \textit{Flat Triplets} and \textit{Nested Hierarchy} are evaluated using the main structure-aware L1-L3 protocol in~\autoref{sec:evaluation_suite}. 
Thus, the ablation isolates output organization: both variants predict the same fields and use the same hierarchical metrics, differing only in whether parent-child bindings are generated natively or reconstructed post hoc.

\section{Downstream Interaction-Oriented Reasoning Probe}
\label{appendix:planning_probe}

We use the same task instruction under two prompting conditions. 
In the task-only setting, the planner receives the RGB image and task prompt, which may specify target object semantics, and is asked to generate step-by-step manipulation plans with manipulated objects, bounding boxes, and interaction points. 
In the structure-augmented setting, the planner additionally receives the SceneParser hierarchy, which links objects to functional parts, affordance labels, and interaction points.

As shown in~\autoref{fig:planning_probe}, task-only prompting can identify some task-relevant objects but may miss functional parts or produce incomplete affordance assignments. 
With the SceneParser hierarchy, the planner more consistently grounds steps to object-part-affordance chains, such as selecting the drawer handle for pulling and the drawer interior for placing. 
This comparison shows that SceneParser adds structured object-part-affordance bindings beyond task-provided object semantics, converting local perception into a decision-ready interface.

\input{Figures/probe}

%% file: Figures/probe.tex

\begin{center}
    \includegraphics[width=0.8\linewidth]{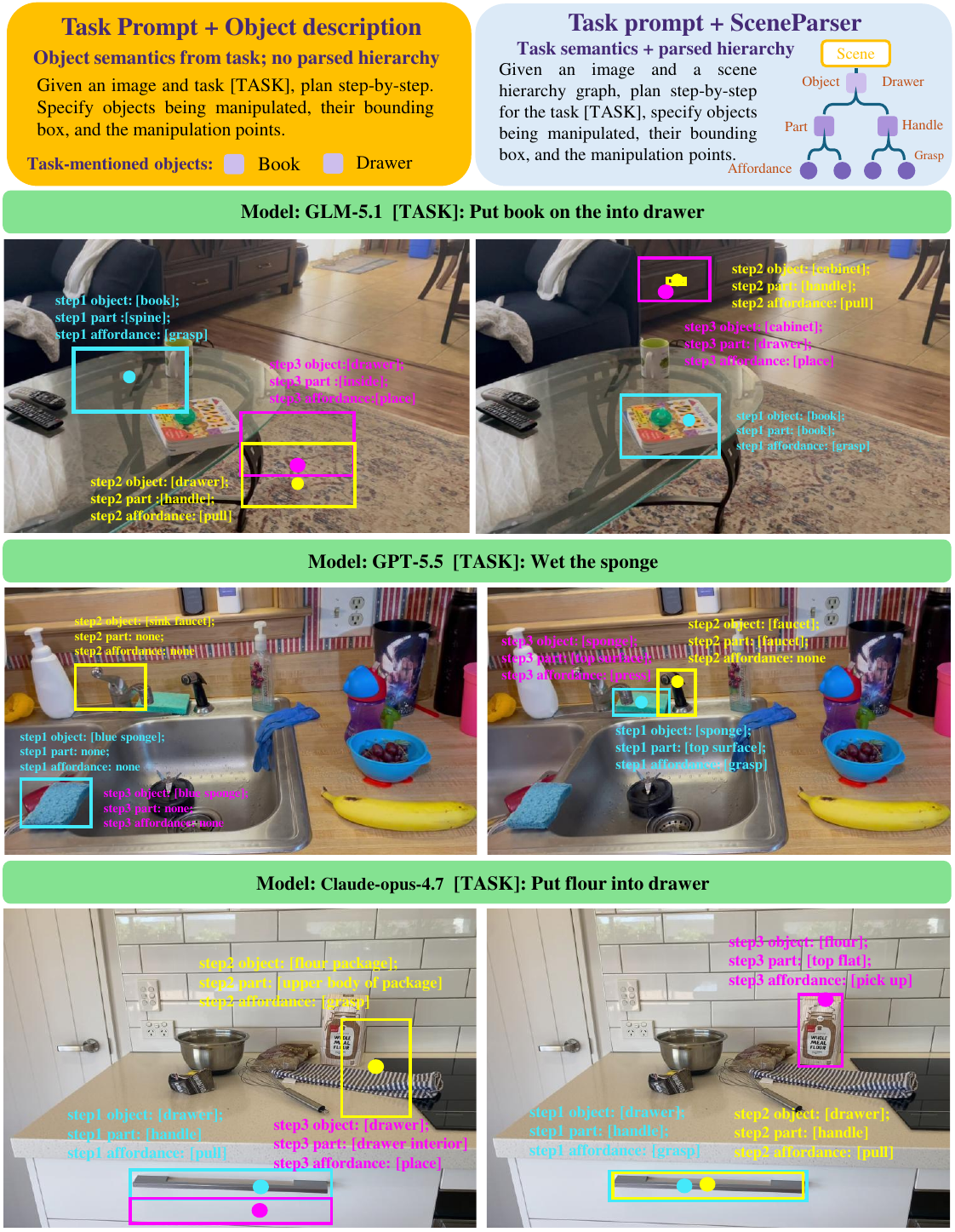}
    \captionof{figure}{
    Qualitative downstream interaction-oriented reasoning probe.
    We compare a task-only prompt, which already provides task-level object semantics, with a structure-augmented prompt that additionally provides the object-part-affordance hierarchy produced by SceneParser.
    The hierarchy exposes functional parts, affordance labels, and interaction points, helping the planner generate more coherent interaction steps and more accurate object-part-affordance grounding.
    }
    \label{fig:planning_probe}
\end{center}